\documentclass{article}

\usepackage{arxiv}

\usepackage[utf8]{inputenc} 
\usepackage[T1]{fontenc}    
\usepackage{hyperref}       
\usepackage{url}            
\usepackage{booktabs}       
\usepackage{amsfonts}       
\usepackage{nicefrac}       
\usepackage{microtype}      
\usepackage{lipsum}		
\usepackage{amsthm}
\usepackage{graphicx}
\usepackage{doi}
\usepackage{physics}
\usepackage{amssymb}
\usepackage{pifont}

\newcommand{\pd}[2]{\frac{\partial #1}{\partial #2}}
\newcommand{\pdd}[2]{\frac{\partial^2 #1}{\partial #2^2}}

\usepackage{setspace}
\title{Nonlinear discretizations and Newton's method: characterizing stationary points of regression objectives}

\date{} 					

\author{
    Conor Rowan \\
	Smead Aerospace Engineering Sciences\\
	University of Colorado Boulder\\
    3775 Discovery Drive\\
	Boulder, CO 80309 \\
	\texttt{conor.rowan@colorado.edu} \\
}

\renewcommand{\headeright}{}
\renewcommand{\undertitle}{}
\renewcommand{\shorttitle}{}

\hypersetup{
pdftitle={A template for the arxiv style},
pdfsubject={q-bio.NC, q-bio.QM},
pdfauthor={David S.~Hippocampus, Elias D.~Striatum},
pdfkeywords={First keyword, Second keyword, More},
}

\begin{document}
\maketitle

\begin{abstract}
	Second-order methods are emerging as promising alternatives to standard first-order optimizers such as gradient descent and ADAM for training neural networks. Though the advantages of including curvature information in computing optimization steps have been celebrated in the scientific machine learning literature, the only second-order methods that have been studied are quasi-Newton, meaning that the Hessian matrix of the objective function is approximated. Though one would expect only to gain from using the true Hessian in place of its approximation, we show that neural network training reliably fails when relying on exact curvature information. The failure modes provide insight both into the geometry of nonlinear discretizations as well as the distribution of stationary points in the loss landscape, leading us to question the conventional wisdom that the loss landscape is replete with local minima.
\end{abstract}



\section{Introduction}

\paragraph{} First-order optimization methods have historically presided over the training of neural networks. These methods find a minimum of the objective function by iteratively updating the neural network parameters in the direction that most rapidly decreases this objective. In the case of ADAM---a widely used first-order algorithm which includes ``momentum"---the update to the parameters is a weighted combination of the current and past steepest descent directions \cite{kingma_adam_2017}. While ADAM has been successful on an impressive array of problems in scientific machine learning (SciML) \cite{raissi_physics-informed_2019, khodayi-mehr_varnet_2019, e_deep_2017, lu_deeponet_2021, li_fourier_2021}, recent years have seen growing interest in second-order optimization methods for physics-informed machine learning problems. Second-order optimization strategies use a local quadratic approximation of the objective function to determine the step direction and step size. One advantage of second-order methods is that the local quadratic approximation---by virtue of containing stationary points---naturally suggests a step size, unlike the unbounded linear approximation underlying steepest-descent methods \cite{ruszczynski_nonlinear_2006}. Another advantage is that the step direction avoids oscillations and slow convergence in ill-conditioned regions of the loss landscape, a well-known shortcoming of first-order methods \cite{nocedal_numerical_2006}. Because of these advantages, second-order optimization methods are already standard in many areas of science and engineering. Accordingly, many authors have begun to investigate these strategies for optimization problems arising from machine learning. In this work, our attention will be limited to regression problems inspired by the SciML literature. In particular, we have in mind physics-informed neural networks (PINNs), for which the target function is defined in space and/or time and the regression problem involves spatial and/or temporal derivatives of the neural network regressor. The basic PINNs approach was first introduced in \cite{sirignano_dgm_2018, raissi_physics-informed_2019}, and has since been explored extensively in many interesting scientific and engineering application areas \cite{manav_phase-field_2024, jin_nsfnets_2021, abueidda_deep_2022, cai_physics-informed_2021}. 

\paragraph{} Before proceeding, we distinguish between exact Newton and quasi-Newton methods for second-order optimization. An exact Newton method uses the matrix of second-derivatives of the objective function---in other words, the Hessian matrix---to build a local quadratic approximation of the objective. The Hessian matrix is either computed analytically and supplied to the optimizer or computed with finite differencing, though the cost of the latter is typically prohibitive. In contrast, a quasi-Newton method constructs a running approximation of the Hessian from observations of the loss and its gradient obtained over the optimization history. At each optimization step, the previous estimate of the Hessian is updated based on new observations. A common update strategy is provided by the method of Broyden, Fletcher, Goldfarb, and Shanno (BFGS), which finds a minimum Frobenius norm update to the previous Hessian subject to the constraint that the update is consistent with a Taylor approximation (secant condition) \cite{nocedal_quasi-newton_1999}. A popular modification of this approach is the limited-memory BFGS (L-BFGS) algorithm, which avoids storing a dense approximation of the inverse Hessian and is thus more efficient for high-dimensional problems. 

\paragraph{} To the best of the author's knowledge, all second-order methods explored in the SciML literature have been quasi-Newton. In \cite{sun_physics-informed_2023}, the authors use PINNs to optimize an airfoil geometry while simultaneously learning the flow distribution with L-BFGS. Because of the ability of second-order quasi-Newton methods to converge rapidly even in ill-conditioned loss landscapes, a novel hybrid of L-BFGS and ADAM is proposed in \cite{rathore_challenges_2024}, demonstrating improved performance for physics-informed objective functions. One study compares optimizers in the BFGS family on challenging PINNs problems, demonstrating that self-scaled variants of BFGS and Broyden optimizers lead to marked performance improvements \cite{kiyani_optimizing_2025}. Another work shows that quasi-Newton methods naturally resolve conflicts between gradients of different terms in the loss function, which often lead to poor convergence for first-order methods \cite{wang_gradient_2025}. The authors in \cite{urban_unveiling_2025} report that minor modifications to the loss function and standard BFGS algorithm allow PINNs solutions to outperform incumbents such as finite difference methods. Finally, in \cite{ahmad_preconditioned_2025}, a novel quasi-Newton preconditioning strategy is developed and shown to outperform state-of-the-art methods such as L-BFGS. These works all convincingly demonstrate the advantages of incorporating second-order curvature information into neural network training. 

\paragraph{} Though quasi-Newton methods for training neural networks have seen a surge of interest in recent years, exact Newton methods remain unexplored. We speculate that the well-known computational cost of analytically forming, storing, and inverting the Hessian for large-scale problems has discouraged such inquiry. However, we argue in this work that there are more fundamental reasons why exact Newton methods are infeasible for machine learning applications. We show through a mix of theoretical analysis and numerical experimentation that exact Newton methods exhibit surprising pathological behaviors on regression problems. We then leverage these failures to gain insight into the geometry of neural network discretizations and the structure of stationary points in the loss landscape. In particular, our contributions are as follows:

\begin{enumerate}
    \item We discuss regression on manifolds, showing that stationary points can be interpreted geometrically;
    \item We conceptualize neural networks as defining an approximation manifold which constructs basis functions and their coefficients simultaneously;
    \item We identify one particular stationary point of the neural network-based regression objective which represents a trivial zero solution to the regression problem;
    \item We show through numerical experimentation that exact Newton methods reliably find these trivial solutions, even for simple one-dimensional problems;
    \item We discuss the ways in which quasi-Newton methods differ from exact Newton methods and how these differences lead to the documented successes of the former.
\end{enumerate}

The rest of this work is organized as follows. In Section 2, we discuss nonlinear discretizations and regression on manifolds. We show that stationary points of the regression objective need not be minima and that the condition for stationarity can be interpreted geometrically. In Section 3, we turn to regression with neural networks, showing that multilayer perceptron neural networks can be viewed as simultaneously fitting and scaling basis functions. Such structure allows for stationarity of the regression objective to be satisfied by zeroing the coefficients of all basis functions, so long as the basis functions satisfy a certain property. We show with numerical examples that neural networks trained with exact Newton methods reliably converge to precisely this trivial solution. We then use the eigenvalues of the Hessian matrix to determine that this solution is a saddle point rather than a local minimum or maximum. In Section 4, we explore two boundary value problems with PINNs, showing that trivial zero solutions are also theoretically possible in the case of physics-informed training. Though the conditions under which such a solution is obtained appear as or more complex than minimizing the error, we again show that exact Newton methods favor these trivial saddle solutions. Finally, in Section 5 we conclude with reflections on high-dimensional loss landscapes and a discussion of how quasi-Newton methods avoid saddle points.


\section{Nonlinear discretizations}

\paragraph{} Consider a discrete regression problem in which a target vector $\mathbf{v}$ is to be approximated by a parameterized vector $\mathbf{N}(\boldsymbol \theta)$, where  $\boldsymbol \theta$ are the parameters to be determined. As is standard, the regression problem is treated with a quadratic error objective and its stationarity condition:

\begin{equation}\label{orth1}
    \mathcal{L}( \boldsymbol \theta) = \lVert  \mathbf{N}(\boldsymbol \theta) - \mathbf{v}\rVert^2, \quad\pd{\mathcal{L}}{\theta_k} = ( \mathbf{N}(\boldsymbol \theta) - \mathbf{v} ) \cdot \pd{\mathbf{N}}{ \theta_k} = 0.
\end{equation}

Eq. \eqref{orth1} shows that stationarity of the quadratic objective enforces that the error vector $\mathbf{e}(\boldsymbol \theta) = \mathbf{N}(\boldsymbol \theta) - \mathbf{v}$ is orthogonal to the tangent space of the approximation given by $\text{span}\{ \pd{\mathbf{N}}{\theta_1} , \pd{\mathbf{N}}{\theta_2},\dots \}$. When the approximation is linear, the parameters scale fixed basis vectors $\{ \mathbf{h}_j\}_{j=1}^{|\boldsymbol \theta|}$. In this case, the condition for stationarity reads

\begin{equation}\label{galerkin}
    \pd{\mathcal{L}}{\theta_k} = ( \sum_{j=1}^{|\boldsymbol \theta|} \theta_j \mathbf{h}_j - \mathbf{v} ) \cdot \mathbf{h}_k = \mathbf{e}(\boldsymbol \theta) \cdot \mathbf{h}_k =  0.
\end{equation}

The satisfaction of Eq. \eqref{galerkin} is known as Galerkin optimality, which states that the error vector has zero projection in the space defined by the basis vectors. If the basis functions are linearly independent, the stiffness matrix $K_{jk} =  \mathbf{h}_j \cdot \mathbf{h}_k $ is both full-rank and positive definite. These conditions ensure that the optimization problem 1) has a unique solution and 2) that this solution is a minimum. In other words, when the approximation is linear, the basis is fixed and the stationary point of Eq. \eqref{orth1} minimizes the error. This ensures that we obtain as good an approximation of the target vector $\mathbf{v}$ as the basis allows. No such guarantees exist for a nonlinear discretization, which is any parameterization $\mathbf{N}( \boldsymbol \theta)$ other than that of a fixed basis with variable coefficients. We view these nonlinear discretizations as defining approximations on manifolds, or surfaces embedded in a space with more dimensions than there are parameters. Though mathematicians may object to this simplistic definition of a manifold, it suffices for our purposes. To illustrate this embedding, consider the following regression problem with a nonlinear discretization:

\begin{equation*}
    \mathbf{N}(\theta) = \begin{bmatrix}
        \cos(\theta) \\ \sin(\theta)
    \end{bmatrix}, \quad \mathbf{v} = \begin{bmatrix}
         2 \\ 2
    \end{bmatrix}, \quad \theta \in[0,2\pi).
\end{equation*}

The dimensionality of the optimization problem is $|\boldsymbol \theta|=1$, yet the approximation lives on the unit circle in $\mathbb{R}^2$. Galerkin optimality requires that the error vector is orthogonal to the tangent of the approximation, which reads

\begin{equation*}
    \pd{\mathcal{L}}{\theta} = \begin{bmatrix}
        \cos(\theta) - 2 \\ \sin(\theta) - 2 
    \end{bmatrix} \cdot \begin{bmatrix}
        - \sin(\theta) \\ \cos(\theta)
    \end{bmatrix} =  2(\sin(\theta) - \cos(\theta) ) = 0.
\end{equation*}

\begin{figure}[hbt!]
\centering
\includegraphics[width=1.0\textwidth]{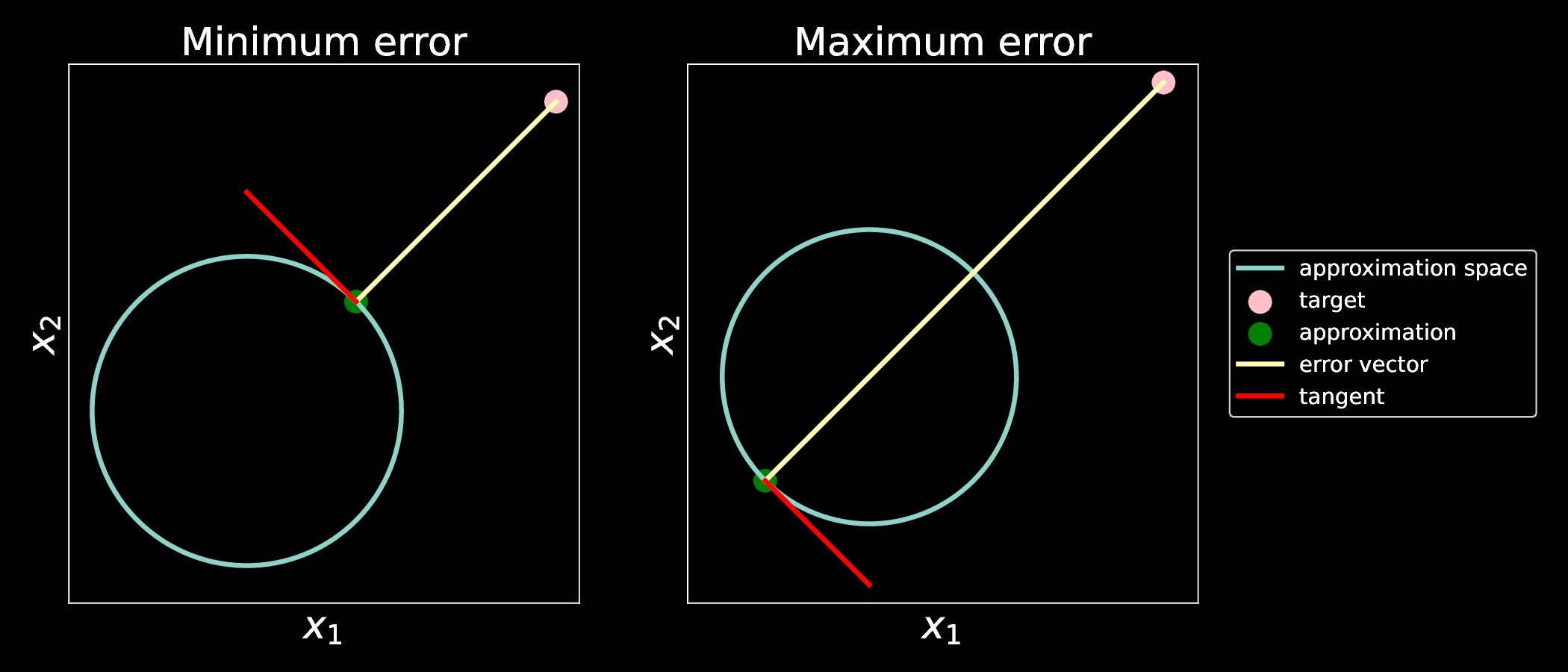}
\caption{The error vector is orthogonal to the tangent of the unit circle approximation space both when the magnitude of the error vector is minimized and when it is maximized. }
\label{circle}
\end{figure}

This equation is satisfied when $\theta =  \pi/4, 5\pi/4$. Evidently, unlike the linear discretization, the uniqueness of the solution is no longer guaranteed. Furthermore, computing the second derivative of the loss as $\partial^2 \mathcal{L} / \partial \theta^2 = 2( \cos(\theta) + \sin(\theta))$, we observe that 

\begin{equation*}
    \pdd{\mathcal{L}}{\theta} \Bigg|_{\pi/4} = 2\sqrt{2} >0, \quad \pdd{\mathcal{L}}{\theta} \Bigg|_{5\pi/4} = -2\sqrt{2} <0,
\end{equation*}

\noindent which indicates that $\pi/4$ corresponds to a minimum and $5\pi/4$ to a maximum. Thus, not only does the nonlinear discretization have multiple solutions, the error vector can be orthogonal to the tangent(s) of the approximation space even when the objective is maximized. Finding a stationary point per Eq. \eqref{orth1} is no longer a trustworthy guide to good performance on regression problems when the discretization is nonlinear. The satisfaction of the orthogonality condition at both the minimum and maximum error is shown in Figure \ref{circle}. We remark that linear discretizations always define hyperplane approximation spaces, where the dimension of the hyperplane is equivalent to the number of parameters. This contrasts with the unit circle approximation space, which is embedded in $\mathbb{R}^2$ but controlled by a single parameter. 

\begin{figure}[hbt!]
\centering
\includegraphics[width=1.0\textwidth]{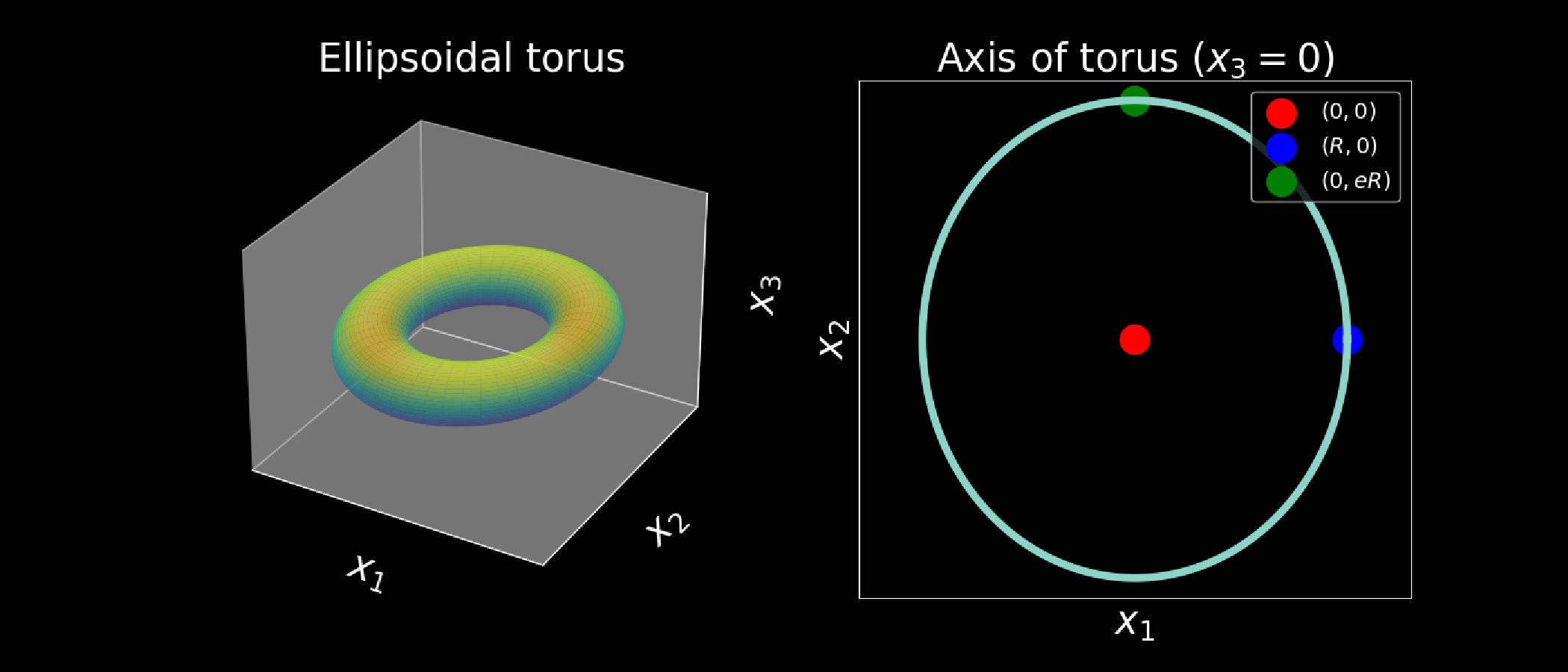}
\caption{The nonlinear discretization of a vector in $\mathbb{R}^3$ is defined by two parameters that traverse the surface of an ellipsoidal torus. The approximation space is visualized in 3D (left) and in cross-section (right).}
\label{torus_fig}
\end{figure}

\paragraph{} When the discretization has only one parameter, the solution to the regression problem is necessarily either a minimum or a maximum. With two or more parameters, it is possible to find stationary points of the loss that are saddle points. A saddle point is a minimum with respect to some parameters and a maximum with respect to others. In order to visualize the geometry of saddle points with nonlinear discretizations, we now define the approximation space as the surface of an ellipsoidal torus. The torus is parameterized by 

\begin{equation*}
    \mathbf{N}(\boldsymbol \theta) = \begin{bmatrix}
        (R + r \cos(\theta_2))\cos(\theta_1) \\ (R+r\cos(\theta_2)) e\sin(\theta_1) \\ r\sin(\theta_2)
    \end{bmatrix}, \quad \theta_1 , \theta_2 \in [0,2\pi),
\end{equation*}

\noindent where $\boldsymbol \theta = [\theta_1,\theta_2]$ are the parameters, $R$ is the radius of the axis of the torus, $r$ controls the thickness of the torus, and $e$ is the eccentricity defining the ratio of the major to the minor axis of the ellipsoidal axis of the torus. See Figure \ref{torus_fig} to visualize the geometry of the approximation space. We look for points on the torus to approximate the origin, i.e., a target vector of $\mathbf{v}=[0,0,0]^T$. The regression problem is thus

\begin{equation}\label{torus}
    \mathcal{L}( \boldsymbol \theta) = \lVert  \mathbf{N}(\boldsymbol \theta) \rVert^2, \quad\pd{\mathcal{L}}{\theta_k} = \mathbf{N}(\boldsymbol \theta) \cdot \pd{\mathbf{N}}{ \theta_k} = 0.
\end{equation}

This stationarity condition states that the position vector is orthogonal to the two tangent vectors. To solve this nonlinear system of equations, we use Newton's method:

\begin{equation}\label{newton}
    \boldsymbol \theta_{k+1} = \boldsymbol \theta_k - \qty( \frac{\partial^2 \mathcal{L}}{\partial \boldsymbol \theta \partial \boldsymbol \theta})^{-1} \Bigg|_{\boldsymbol \theta_k} \qty(\pd{\mathcal{L}}{\boldsymbol \theta}) \Bigg |_{\boldsymbol \theta_k},
\end{equation}

\noindent where we define the Hessian matrix as $\mathbf{J} := \partial^2 \mathcal{L} / \partial \boldsymbol \theta \partial \boldsymbol \theta$ \cite{ruszczynski_nonlinear_2006}. The Newton iterations continue until $\lVert \partial \mathcal{L} / \partial \boldsymbol \theta \rVert \approx 0$, meaning that the nonlinear system of equations for stationarity has been approximately solved. All first and second parameter derivatives are computed with automatic differentiation in PyTorch. 

\begin{figure}[hbt!]
\centering
\includegraphics[width=1.0\textwidth]{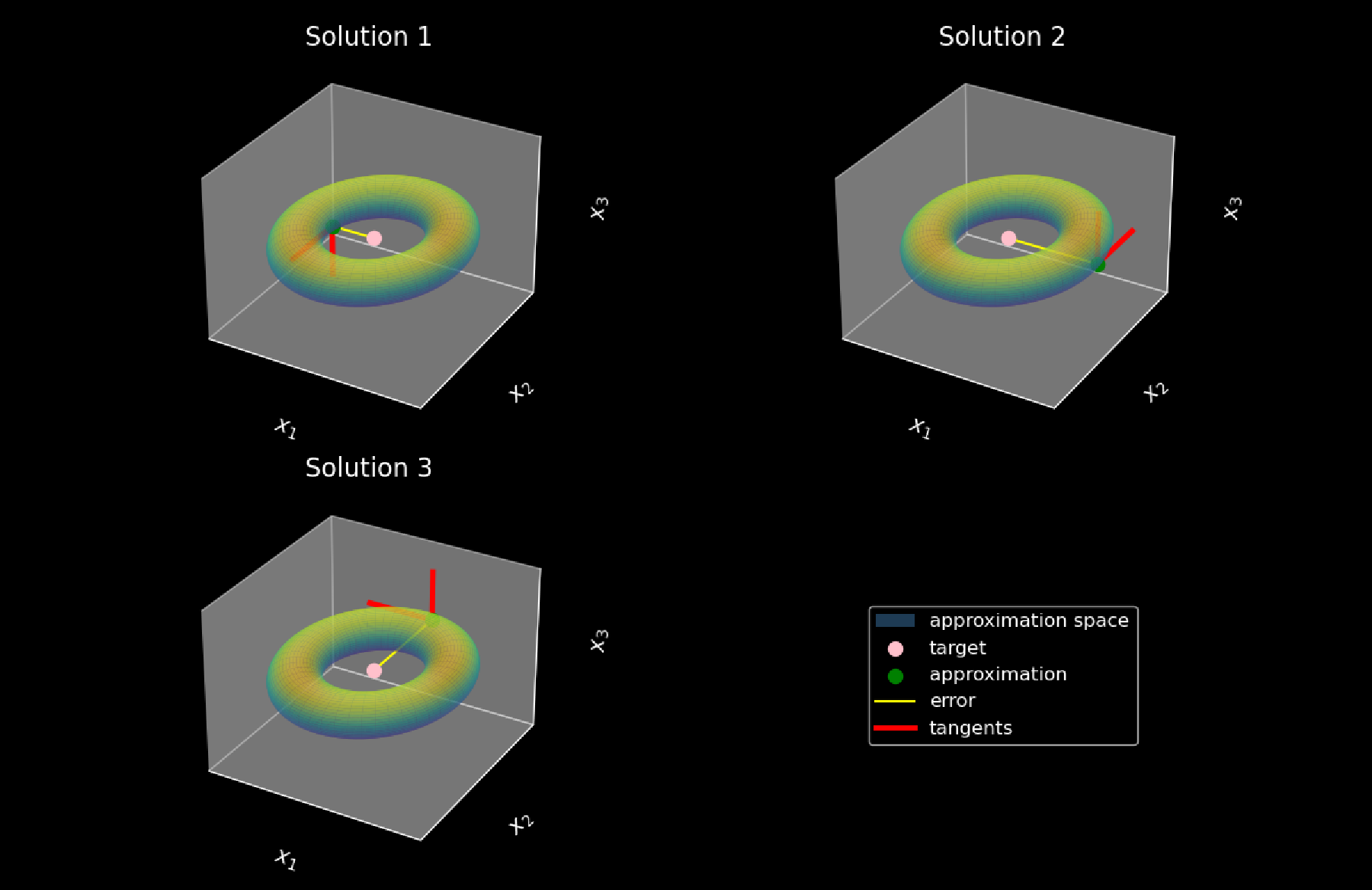}
\caption{There are multiple stationary points of the objective function for the regression problem. Using the Hessian matrices, we classify Solution 1 as a minimum, Solution 2 as a saddle point, and Solution 3 as a maximum.}
\label{solz}
\end{figure}

\paragraph{} In this example, the initial guess of each parameter is independent and uniformly distributed in $[0,2\pi)$. The geometric properties of the torus are given by $R=1$, $r=0.35$, and $e=1.2$. See Figure \ref{solz} for the results of the Newton solution for three different initializations of the parameters. In each of the three runs, we obtain a different point on the torus that satisfies stationarity. As before, it is clear geometrically that the error vector is orthogonal to the tangent space of the approximation, yet only one of these three solutions minimizes the error. To understand the nature of the stationary points, we turn to the Hessian matrix $\mathbf{J}$ whose eigenvalues dictate whether the solution is a minimum, maximum, or saddle. Because the Hessians at the three solutions are diagonal, this allows us to read off the eigenvalues directly as the diagonal entries. The three Hessian matrices are:

\begin{equation*}
    \mathbf{J}_1 = \begin{bmatrix}
        3.7 & 0 \\ 0 & 0.7
    \end{bmatrix}, \quad  \mathbf{J}_2 = \begin{bmatrix}
        1.6 & 0 \\ 0 & -0.7
    \end{bmatrix}, \quad  \mathbf{J}_3 = \begin{bmatrix}
        -1.6 & 0 \\ 0 & -1.1
    \end{bmatrix}.
\end{equation*}

As seen by the sign of the diagonal entries, Solution 1 is a minimum, Solution 2 is a saddle point, and Solution 3 is a maximum. The eigenvalues report how the distance from the origin changes as we traverse the surface of the torus in each of the eigenvector directions, which in this case align with the tangent directions. A positive eigenvalue means that the loss is locally convex in the eigenvector/tangent direction. A negative eigenvalue indicates concavity. In the case of Solution 2, it is interesting to consider how the elliptical shape of the $x_3$ cross-sections of the torus ensures that moving along the torus with the tangent in the $x_3$ direction decreases the distance from the origin whereas moving with the $x_2$ tangent increases the distance. Such mixed curvature is the definition of a saddle point.

\begin{figure}[hbt!]
\centering
\includegraphics[width=1.0\textwidth]{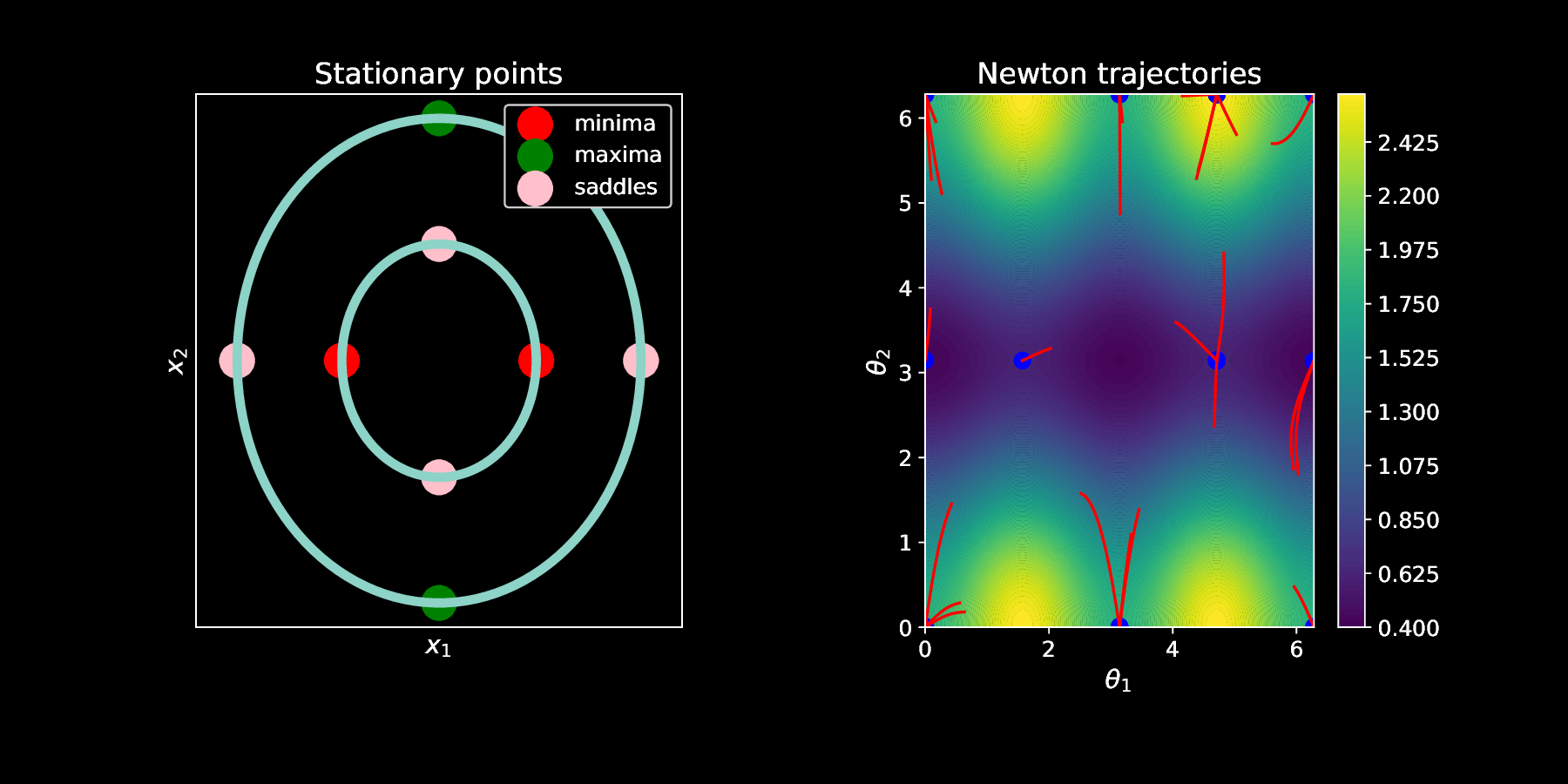}
\caption{All stationary points are in the $x_3=0$ plane and lie along one of the coordinate axes (left). We show $25$ convergence histories of Newton's method for random initializations of the parameters (right). Converged solutions are indicated by blue dots. Note that by periodicity, the saddle points at each of the four corners are actually the same solution. This is also the case for the minimum found along the left and right edge of the domain and the saddle at the center of the top and bottom edges.}
\label{pts}
\end{figure}

\paragraph{} In total, there are $8$ stationary points of the regression objective given in Eq. \eqref{torus}. With the geometric interpretation of stationarity in mind, it is straightforward to see that there are $8$ positions on the torus where the position vector is orthogonal to the tangent space. All stationary points lie in the $x_3=0$ plane. See Figure \ref{pts} for a visualization and characterization of each of the stationary points, as well as a depiction of the dynamics of Newton optimization in the loss landscape. There are $2$ minima, $2$ maxima, and $4$ saddle points. The convergence trajectories indicate that optimization based on an exact Newton method has no preference for minima over other stationary points. This is a well-known consequence of Newton's method looking for a zero of the gradient, rather than a minimum of the objective.

\section{Regression with neural networks}

\paragraph{} One of the most widely used nonlinear discretizations is that of a neural network. We now leverage the intuition acquired from the unit circle and torus examples to interpret Newton solutions to a standard neural network regression problem. We take the architecture as a multilayer perceptron (MLP) neural network. In an MLP, the input-output relation for the $i$-th hidden layer is

\begin{equation*}
    \mathbf{y}_i = \sigma\Big(  \mathbf{W}_i\mathbf{y}_{i-1} + \mathbf{B}_i  \Big),
\end{equation*}

\noindent where $\sigma(\cdot)$ is a nonlinear activation function applied element-wise. As shown, the output $\mathbf{y}_i$ then becomes the next layer's input. The parameters of the neural network are the collection of the weight matrices $\mathbf{W}_i$ and bias vectors $\mathbf{B}_i$ for each layer. Thus, we can write the neural network parameters as $\boldsymbol{\theta} = [ \mathbf{W}_1, \mathbf{B}_1, \mathbf{W}_2,\mathbf{B}_2,\dots]$. Typically, a linear mapping with no bias is used to go from the last hidden layer to the output. 

\paragraph{} A continuous regression problem is defined with a scalar target function $v(x)$ and a neural network discretization $\mathcal{N}(x; \boldsymbol \theta)$ where $x \in [0,1]$ and $\boldsymbol \theta$ is the collection of trainable parameters. The objective function is

\begin{equation}\label{nnregression}
    \mathcal{L}(\boldsymbol \theta) = \frac{1}{2} \int_0^1 \Big(  \mathcal{N}(x;\boldsymbol \theta) - v(x) \Big)^2 dx.
\end{equation}

Using Newton's method, we find a zero of the following nonlinear system of equations corresponding to stationarity of the objective:

\begin{equation}\label{nnstationary}
    \pd{\mathcal{L}}{\boldsymbol \theta} = \int_0^1 \Big(  \mathcal{N}(x;\boldsymbol \theta) - v(x) \Big) \pd{\mathcal{N}}{\boldsymbol \theta} dx
\end{equation}

The connection to the finite-dimensional stationary condition is made clear when Eq. \eqref{nnstationary} is numerically integrated. Approximating the integral on a uniform grid, this reads

\begin{equation*}
    \pd{\mathcal{L}}{\theta_j} = \Delta x \sum_{i=1}^N \Big( \mathcal{N}(x_i ; \boldsymbol \theta) - v(x_i) \Big) \pd{\mathcal{N}(x_i, \boldsymbol \theta)}{\theta_j}  = \Delta x \mathbf{e}(\boldsymbol \theta) \cdot \pd{\mathbf{N}}{\theta_j} = 0, \quad j=1,2,\dots
\end{equation*}

\noindent where $\{ x_i \}_{i=1}^N$ are the integration points. Even in the continuous setting, stationarity of a quadratic loss dictates that the error is orthogonal to the tangent space of the approximation. With this in mind, we show that the structure of MLP neural networks gives certain stationary points a clear interpretation. First, we write the neural network discretization in a more revealing form as

\begin{equation}\label{basis}
    \mathcal{N}( x , \boldsymbol \theta ) = \sum_{k=1}^{|\boldsymbol \theta^{\text{O}}|} \theta^{\text{O}}_k h_k(x; \boldsymbol \theta^{\text{I}}),
\end{equation}

\noindent where the total parameter set $\boldsymbol \theta = [ \boldsymbol \theta^{\text{I}} , \boldsymbol \theta^{\text{O}}]$ is decomposed into an ``inner" ($I$) and ``outer" ($O$) part. The inner parameters define the last layer of the network, which we interpret as basis functions $h_k(x;\boldsymbol \theta^{\text{I}})$, and the outer parameters act as coefficients scaling these basis functions. Note that this assumes there is no bias in the output layer of the network. Arguably, the most intuitive solution to Eq. \eqref{nnstationary} is obtained when the discretized error function is zero, given that the zero vector is orthogonal to any tangent space. But as the examples of the circle and torus suggest, nonlinear discretizations harbor stationary points of the regression objective, which are not minima. To see an example of this, let us make the ansatz that $\mathcal{N}(x; \boldsymbol \theta)=0$ is a solution to Eq. \eqref{nnstationary}. In this case, the stationarity condition becomes

\begin{equation}\label{io}
    \pd{\mathcal{L}}{ \boldsymbol \theta} = \int_0^1 v(x) \pd{\mathcal{N}}{\boldsymbol \theta} dx = \begin{bmatrix}
        \int_0^1 v(x) \pd{\mathcal{N}}{\boldsymbol \theta^\text{O}} dx\\ \int_0^1 v(x) \pd{\mathcal{N}}{\boldsymbol \theta^\text{I}} dx
    \end{bmatrix} = \mathbf{0},
\end{equation}

\noindent where we have decomposed the parameter gradient defining the tangent vectors into inner and outer components. Referring to Eq. \eqref{basis}, it is clear that the outer parameter tangents are simply the basis functions at the current setting of the inner parameters:

\begin{equation*}
     \pd{\mathcal{N}}{\theta^\text{O}_j} = h_j(x;\boldsymbol \theta^{\text{I}} ).
\end{equation*}

Thus, with the trivial regression solution $\mathcal{N}=0$, orthogonality of the error to the outer parameter gradients can be satisfied by fitting basis functions that are orthogonal to the target function $v(x)$. But Eq. \eqref{io} shows that in order to satisfy stationarity generally, we also require orthogonality with respect to the inner parameter gradients. The inner parameter gradients can be written as 

\begin{equation*}
    \pd{\mathcal{N}}{\theta^{\text{I}}_{\ell}} = \sum_{k=1}^{|\boldsymbol \theta^{\text{O}}|} \theta^{\text{O}}_k \pd{h_k(x)}{\theta^{\text{I}}_{\ell}}.
\end{equation*}

Given that all functions $v(x)$ are trivially orthogonal to the zero function, we simply require that $\boldsymbol \theta^{\text{O}} = \mathbf{0}$ in order to satisfy orthogonality with respect to the inner parameter tangents. Thus, stationarity of the loss can be obtained by fitting zero coefficients on a basis that is orthogonal to the target function. At this point, we do not claim that neural networks trained with Newton's method actually find this solution. We only show that it is possible in principle to obtain trivial solutions to regression problems when using MLP neural networks trained with Newton's method. Note that such pathological behavior is made possible by the structure of the output layer of MLP neural networks, which gives a linear combination of a customizable basis. There is no analogue of setting the coefficients $\boldsymbol \theta^{\text{O}}=\mathbf{0}$ in order to zero a subset of the tangents in the torus discretization. The tangent vectors of the torus are always nonzero, ensuring that orthogonality of the error cannot be trivially satisfied.

\paragraph{} We now explore which solutions exact Newton optimization finds to a neural network-based regression problem. As noted previously, all second-order optimization studies in the literature have been with quasi-Newton methods to the best of the author's knowledge. To interpret our results, we define the orthogonality of the basis functions with the target function over the course of training as

\begin{equation*}
    O_j(t) = \int_0^1 \hat v(x) \hat h_j(x; \boldsymbol \theta^{\text{I}}(t)) dx,
\end{equation*}

\noindent where $t$ is a pseudo-time variable indicating the optimization epochs, $\hat v(x)$ is the target function normalized to set the integral of its square to unity, and $\hat h_j$ is the $j$-th basis function normalized in the same way. The objective is that of the regression problem given in Eq. \eqref{nnregression}, and the inner and outer gradient magnitudes are computed as $ \lVert  \partial \mathcal{L} / \partial \boldsymbol \theta^{\text{I}}\rVert^2 / | \boldsymbol \theta^{\text{I}}|$ and $ \lVert  \partial \mathcal{L} / \partial \boldsymbol \theta^{\text{O}}\rVert^2 / | \boldsymbol \theta^{\text{O}}|$ respectively. The standard Newton optimization is modified per

\begin{equation}\label{lm}
    \boldsymbol \theta_{k+1} = \boldsymbol \theta_k - \eta\qty( \frac{\partial^2 \mathcal{L}}{\partial \boldsymbol \theta \partial \boldsymbol \theta} + \epsilon \mathbf{I})^{-1} \Bigg|_{\boldsymbol \theta_k} \qty(\pd{\mathcal{L}}{\boldsymbol \theta}) \Bigg |_{\boldsymbol \theta_k},
\end{equation}

\noindent where $0<\eta<1$ relaxes the step size and $\epsilon>0$ introduces convexity into the quadratic approximation of the loss to avoid excessively large steps when the loss has little curvature. This introduction of convexity is known as the Levenberg-Marquardt algorithm \cite{marquardt_algorithm_1963}. Convergence of the Newton updating will halt once $ \lVert \partial \mathcal{L} / \partial \boldsymbol \theta \rVert <\mathcal{T}$ where $\mathcal{T}$ is a problem-specific convergence criterion.

\begin{figure}[hbt!]
\centering
\includegraphics[width=1.0\textwidth]{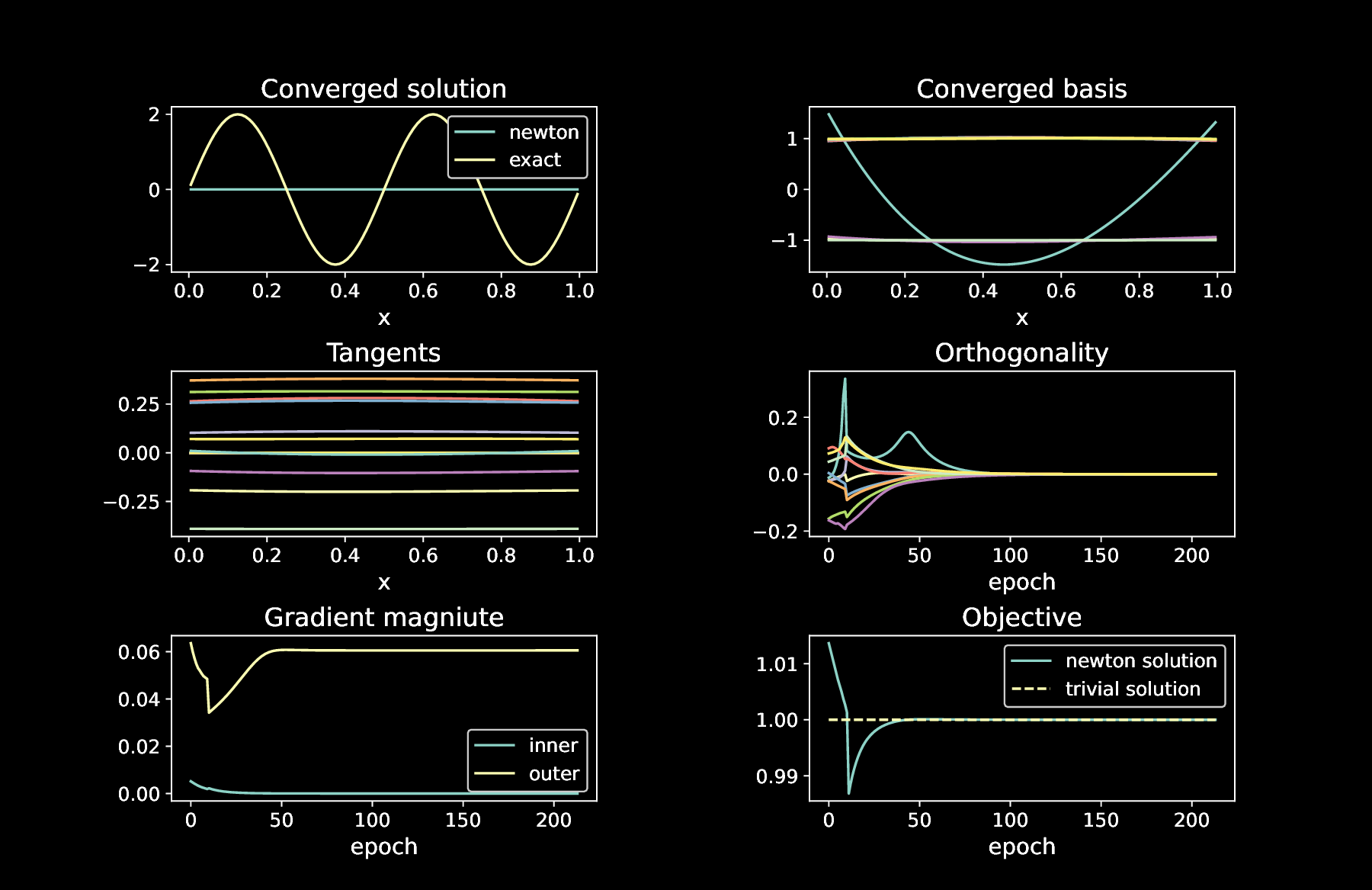}
\caption{Exact Newton optimization obtains the trivial solution we identified. Note that the magnitude of each basis function is normalized to unity for visualization purposes. }
\label{N4}
\end{figure}

\begin{figure}[hbt!]
\centering
\includegraphics[width=0.45\textwidth]{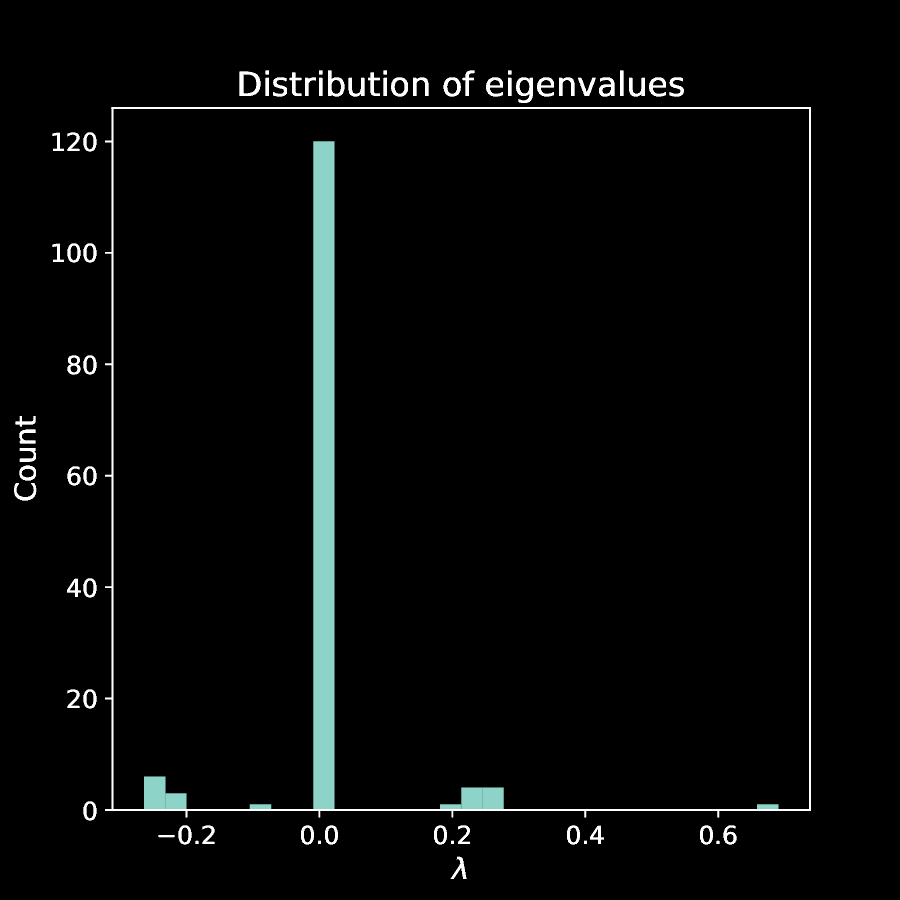}
\caption{Most eigenvalues of the Hessian matrix at the converged solution cluster around zero, indicating that the loss landscape has little to no curvature in the corresponding eigenvector directions. The remaining eigenvalues are an even mix of positive and negative, indicating a saddle point solution. We note that a similar clustering of eigenvalues around zero is observed over the course of training as well, which explains why the Levenberg-Marquardt modification in Eq. \eqref{lm} is required to stabilize training.}
\label{eigN4}
\end{figure}

\paragraph{} In the regression problem, we use a standard MLP architecture with two hidden layers, hyperbolic tangent activation functions, and $10$ neurons per hidden layer with an output layer containing no bias. This corresponds to $| \boldsymbol \theta| = 140$ trainable parameters. Going forward, all neural network parameters will be initialized with the built-in Xavier initialization in PyTorch. We use an integration grid of $100$ equally spaced points. The target function is $v(x) = 2\sin( 4 \pi x)$. In this example, we take $\eta=\epsilon=5 \times 10^{-2}$ and $\mathcal{T} = 1 \times 10^{-5}$. See Figure \ref{N4} for the results of the Newton optimization. Exactly as discussed above, we obtain a trivial solution by finding a basis which is orthogonal to the target function and setting the coefficients to zero. Because the constant function is orthogonal to $\sin(4\pi x)$, all but one of the basis functions is constant. The one non-constant basis resembles $\sin(\pi x) + c$ where $c$ is a constant shift, which is also orthogonal to the target. Figure \ref{eigN4} shows the distribution of eigenvalues of the Hessian at the converged solution. Most eigenvalues are approximately zero, and the remainder represent an approximately even split between positive and negative, indicating that the solution is a saddle point in this high-dimensional loss landscape. 

\paragraph{} We re-run the problem with different initializations to investigate the robustness of this trivial solution. In one numerical experiment, we obtain the trivial solution $9$ out of $10$ runs, though the non-constant basis functions is not always replicated. Note that we verify that the network is capable of accurately representing the target function by solving the regression problem with ADAM optimization, for which convergence to saddle points is not an issue.

\begin{figure}[hbt!]
\centering
\includegraphics[width=1.0\textwidth]{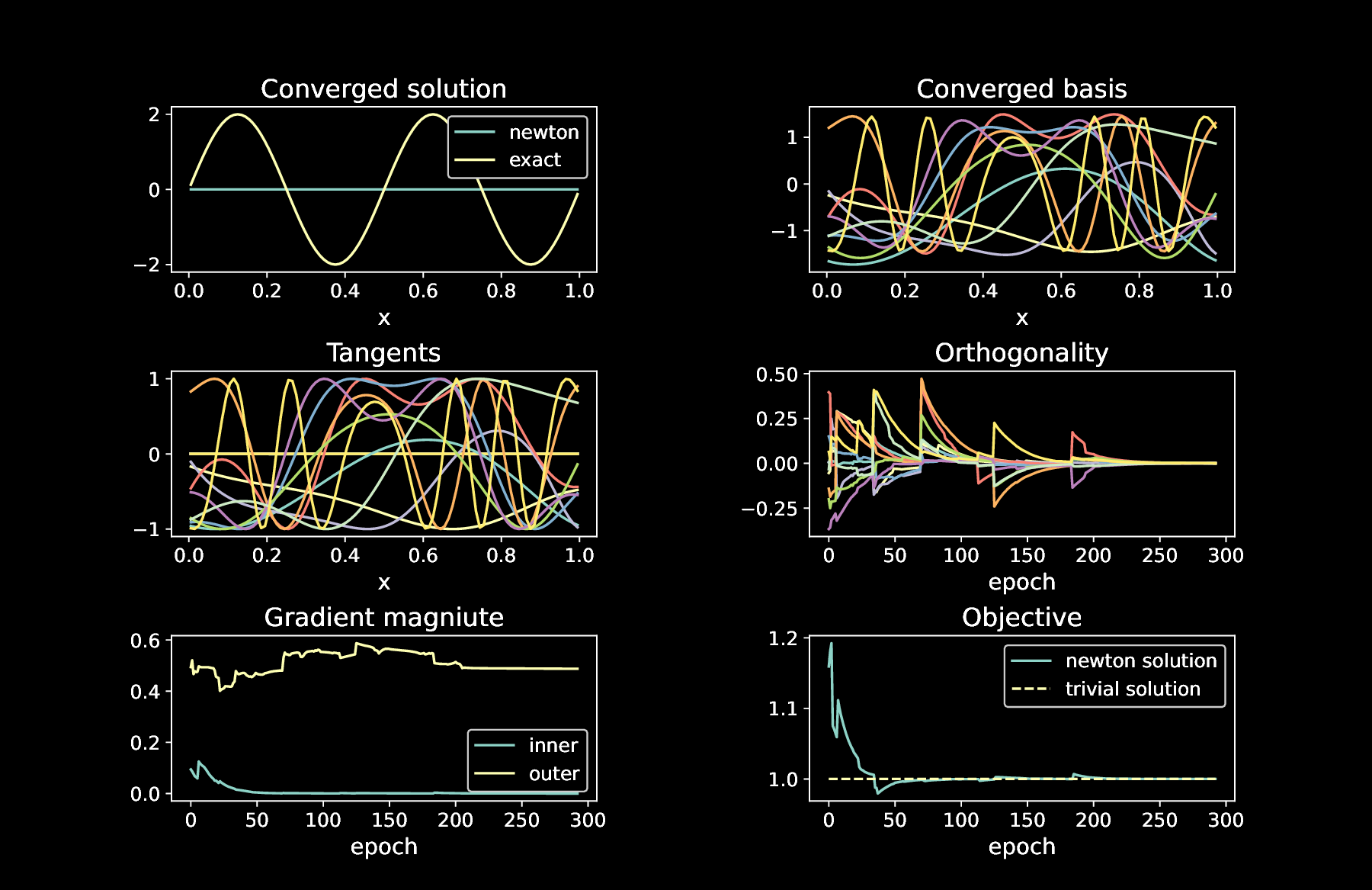}
\caption{The SIREN network manually introduces high-frequency behavior into the neural network, which leads to higher frequency basis functions but does not avoid the trivial saddle point solution.}
\label{Nsin4}
\end{figure}

\paragraph{} It is both interesting and surprising that the neural network reliably converges to a trivial solution by learning an orthogonal basis rather than minimizing the error with the target function. Figure \ref{N4} shows that among all choices of functions orthogonal to $\sin(4\pi x)$, the network favors the two lowest frequency options. We posit that this is a case of the well-known spectral bias of neural networks, which states that standard neural network architectures converge most rapidly on low frequency functions \cite{rahaman_spectral_2019}. In order to push our findings a step further, we manually inject high-frequency behavior into the network by using sinusoidal activation functions, which is known as the sinusoidal representation network (SIREN) \cite{sitzmann_implicit_2020}. We use the same two hidden layer network but replace hyperbolic tangent activation functions with $\sin(\omega_0(\cdot))$ where $\omega_0$ is a hyperparameter controlling the frequency content of the neural network approximation. We use the same target function and the same value of $\eta=5 \times 10^{-2}$ but increase the convexity parameter to $\epsilon=1 \times 10^{-1}$ and the convergence criterion to $\mathcal{T} = 1 \times 10^{-3}$. The frequency hyperparameter is set at $\omega_0=4$. These adjustments account for the more complex loss landscape arising from the oscillatory activation functions. See Figure \ref{Nsin4} for the results of the Newton optimization. Once again, we obtain a trivial solution to the regression problem. However, the basis functions are now 1) high-frequency and 2) non-redundant. As before, the eigenvalues of the Hessian indicate that this solution is a saddle point. \textit{Learning a high-frequency orthogonal basis as opposed to fitting the target function represents a spectacular failure of exact Newton optimization.} Fitting an orthogonal basis of the sort shown in Figure \ref{Nsin4} is in many ways a more complex problem than driving the error to zero by matching the target function. In fact, \cite{rowan_solving_2025} shows that orthogonality of a set of functions is a prohibitively complex optimization objective for neural networks, though this result was obtained without high-frequency behavior injected into the network. In spite of the complexity of explicitly learning an orthogonal basis, one numerical experiment indicated that the trivial solution was obtained $4$ out of $5$ runs of the Newton optimization.

\begin{figure}[hbt!]
\centering
\includegraphics[width=1.0\textwidth]{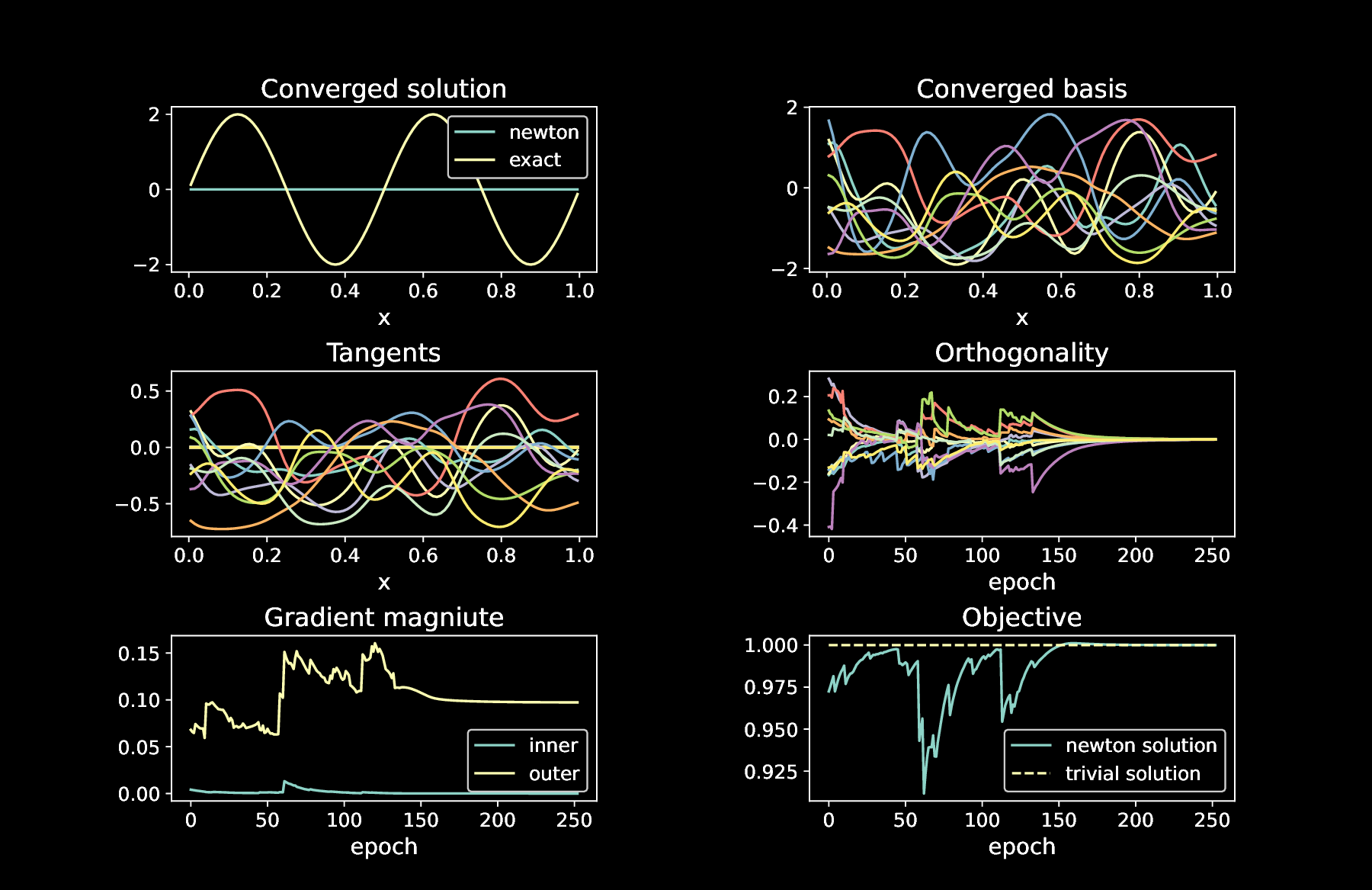}
\caption{The MLP network with $2f=20$ embedded Fourier features also finds a trivial solution to the regression problem with Newton optimization, in spite of the high-frequency basis.}
\label{Nf4}
\end{figure}

\paragraph{} Another strategy for introducing high-frequency behavior into MLP neural networks is that of Fourier feature embedding \cite{wang_eigenvector_2021}. Whereas the standard MLP network takes in the spatial coordinate $x$, the Fourier feature network's input layer is 

\begin{equation*}
    \boldsymbol \gamma(x) = [ \sin( 2\pi \mathbf{B} x) , \cos(2\pi \mathbf{B} x)]^T,
\end{equation*}

\noindent where $\mathbf{B} \in \mathbb{R}^{ f }$ is a vector whose components are normally distributed with variance $\sigma^2$ and $2f$ is the number of Fourier features. The variance of the components of the random embedding matrix $\mathbf{B}$ determines the frequency content of the discretization. Taking $\sigma^2=1.5$ and $f=10$, we use a two hidden layer MLP with hyperbolic tangent activation functions with Fourier features at the input layer. The width of the hidden layers is again $10$, and we set $\eta=5 \times 10^{-2}$, $\epsilon=1 \times 10^{-1}$, and $\mathcal{T} = 1 \times 10^{-3}$. See Figure \ref{Nf4} for the results of the Newton optimization. In spite of the high-frequency basis, we again converge to the trivial solution. Like the SIREN network, the basis functions are high-frequency and non-redundant. From our experience, the trivial solution is obtained with the Fourier features in approximately half of all runs, indicating again that the Newton optimization routinely fails to solve the desired regression problem. The majority of the remaining runs fail to converge at all.

\section{Physics-informed machine learning}

\paragraph{} We now investigate whether a trivial solution analogous to the regression problems of the previous section is obtained in the case of physics-informed training. As a first test, we solve the second-order elliptic boundary value problem given by 

\begin{equation}\label{bvp}
    \pdd{u}{x} + v(x) = 0, \quad u(0)=u(1)=0.
\end{equation}

Per the standard PINNs approach \cite{raissi_physics-informed_2019}, we take the regression objective and modify it to minimize the strong form loss of the governing differential equation:

\begin{equation}\label{physics}
    \mathcal{L}(\boldsymbol \theta) = \frac{1}{2}\int_0^1 \qty( \pdd{\mathcal{N}(x;\boldsymbol \theta)}{x} + v(x))^2 dx, \quad \mathcal{N}(0; \boldsymbol \theta) = \mathcal{N}(1; \boldsymbol \theta)= 0.
\end{equation}

We choose to build the boundary conditions into the neural network discretization of the solution with the distance function method. A distance function discretization for homogeneous Dirichlet boundaries is given by $\mathcal{N}(x; \boldsymbol \theta) = \sin(\pi x) \mathcal{\tilde N}(x;\boldsymbol \theta)$, where $\mathcal{\tilde N}$ is a neural network that need not satisfy the Dirichlet boundary conditions \cite{wang_exact_2023, sukumar_exact_2022, sheng_pfnn_2021}. We modify this distance function approach in order to preserve the interpretation of the inner and outer parameters of the network as building and scaling basis functions. As such, we use the distance function to enforce the homogeneous Dirichlet boundaries at the level of the basis functions themselves. The neural network discretization then reads

\begin{equation*}
    \mathcal{N}(x;\boldsymbol \theta) = \sum_{k=1}^{|\boldsymbol \theta^{\text{O}}|} \theta^{\text{O}}_k \sin(\pi x)  \tilde h_k(x; \boldsymbol \theta^{\text{I}}),
\end{equation*}

\noindent where $\tilde h_k$ represents the $k$-th neuron in the final layer of the network. The basis functions are then $h_k(x;\boldsymbol \theta^{\text{I}}) = \sin(\pi x) \tilde h_k(x; \boldsymbol \theta^{\text{I}})$. This ensures that the solution satisfies the homogeneous Dirichlet boundaries automatically without complicating our breakdown of the neural network into inner and outer parameters. Taking a gradient of Eq. \eqref{physics}, stationarity of the physics objective is given by 

\begin{equation*}
    \pd{\mathcal{L}}{\theta_k} = \int_0^1 \qty( \pdd{\mathcal{N}(x;\boldsymbol \theta)}{x} + v(x)) \frac{\partial^3 \mathcal{N}}{ \partial \theta_k \partial x^2} dx = 0.
\end{equation*}

\begin{figure}[hbt!]
\centering
\includegraphics[width=1.0\textwidth]{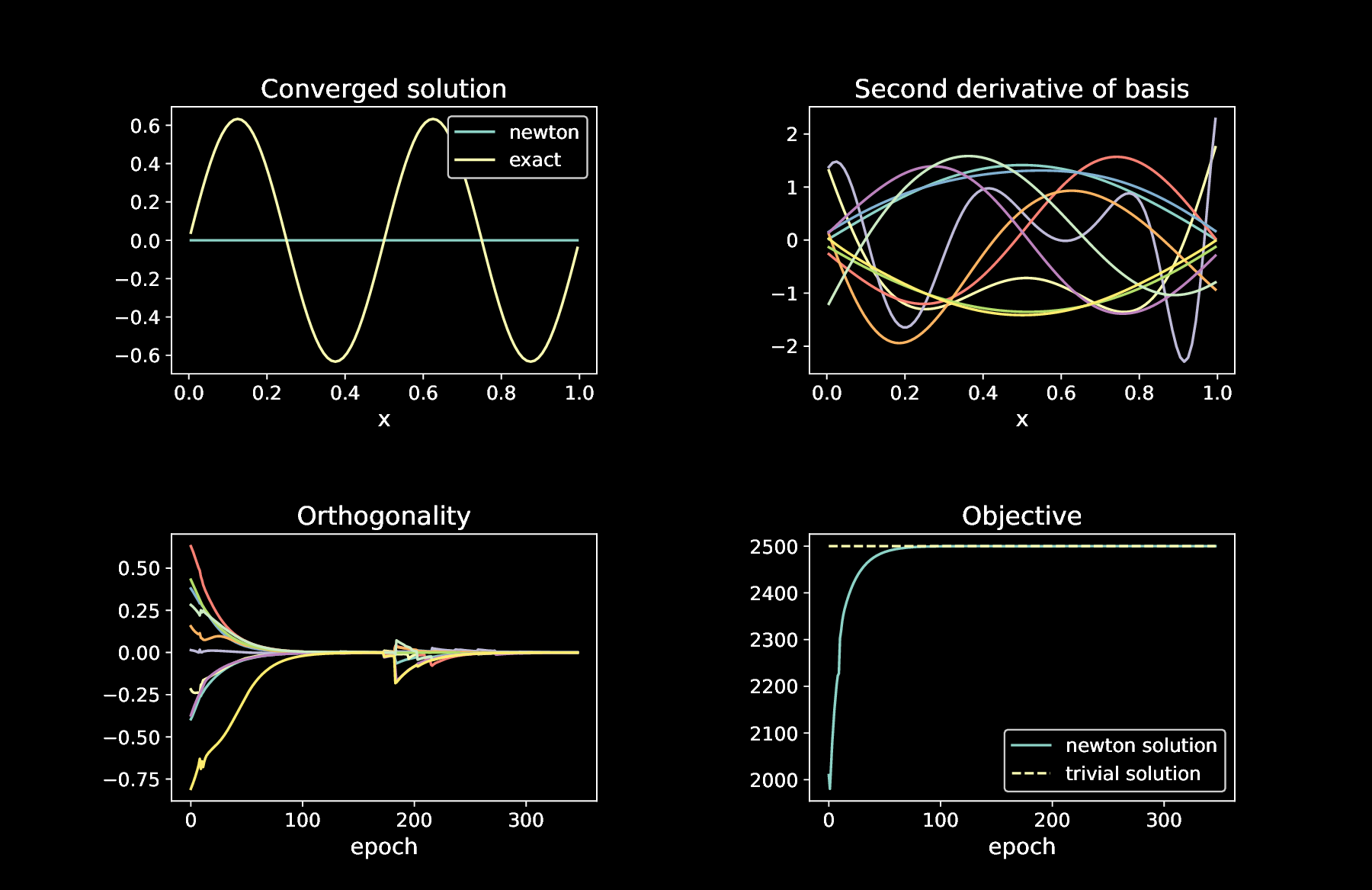}
\caption{Training the network with a physics-based objective does not prevent convergence to trivial solutions. Understanding the nature of these saddle point solutions in the physics-informed regime only requires changing the orthogonality condition to involve the image of the basis under the differential operator.}
\label{pinn_update}
\end{figure}

In order to obtain a trivial solution in the case of PINNs, it is necessary that $\mathcal{N}=0$ and that the forcing function $v(x)$ is orthogonal to second spatial derivatives of each basis function. The relevant orthogonality condition tracked during Newton iteration becomes

\begin{equation*}
    O_j(t) =  \int_0^1  \hat v(x) \frac{\partial^2 \hat h_j(x; \boldsymbol \theta^{\text{I}})}{ \ \partial x^2} dx,
\end{equation*}

\noindent where hats again indicate normalized quantities. We take $v(x) = 100 \sin( 4 \pi x)$ and discretize the solution with a two hidden layer SIREN network of width $10$ and $\omega_0=4$. Owing to the observed two order of magnitude increase in the scale of the objective, we take the convergence criterion to be $\mathcal{T} = 1$ and perform relaxed Newton optimization with $\eta=5 \times 10^{-2}$ and $\epsilon = 1 \times 10^{-1}$. Spatial gradients as well as parameter gradients are now computed with automatic differentiation using PyTorch. See Figure \ref{pinn_update} for the results. Once again, Newton's method converges to a trivial saddle solution. The interpretation of the trivial solution in the case of the PINNs solution is a simple analogue of the standard regression problem: the network learns basis functions whose image under the differential operator are orthogonal to the target function. Note that the target function is now the source term in the boundary value problem of Eq. \eqref{bvp}. Fitting such a basis trivially enforces the stationarity of the physics loss. Like the regression examples, convergence to this solution is robust, occurring $5$ out of $5$ runs in one numerical experiment.

\paragraph{} Thus far, all tests have been in one spatial dimension. As a final exploration of saddle point solutions with neural network discretizations, we solve a diffusion-reaction boundary value problem in two dimensions with homogeneous Dirichlet boundary conditions. The governing equation is given by 

\begin{equation*}
    \begin{aligned}
        \nabla^2 u(\mathbf{x}) + u(\mathbf{x}) + v(\mathbf{x}) = 0, \quad \mathbf{x} \in \Omega,\\
        u(\mathbf{x}) = 0, \quad \mathbf{x} \in \partial \Omega,
    \end{aligned}
\end{equation*}

\noindent where the domain is $\Omega=[0,1]^2$. Using the same modification of the distance function method, the boundary conditions are built into the basis functions:

\begin{equation*}
    \mathcal{N}(\mathbf{x};\boldsymbol \theta) = \sum_{k=1}^{|\boldsymbol \theta^{\text{O}}|} \theta^{\text{O}}_k \sin(\pi x_1) \sin(\pi x_2) \tilde h_k(\mathbf{x}; \boldsymbol \theta^{\text{I}}).
\end{equation*}

With the boundary conditions satisfied automatically by the discretization, the strong form loss is

\begin{equation*}
    \mathcal{L}(\boldsymbol \theta) = \frac{1}{2}\int_{\Omega} \qty( \nabla^2 \mathcal{N}(\mathbf{x};\boldsymbol \theta) + \mathcal{N}(\mathbf{x}; \boldsymbol \theta) +v(\mathbf{x}))^2 d\Omega.
\end{equation*}

Analogous to the one-dimensional PINNs problems, the condition for stationarity of the objective states that the residual is orthogonal to the differential operator applied to the discretization. In the case of the trivial solution, this reads


\begin{equation}\label{orth2d}
    \pd{\mathcal{L}}{\theta_k} = \int_{\Omega}  v(\mathbf{x}) \qty( \frac{\partial^3 \mathcal{N}}{  \partial x_i \partial x_i \partial \theta_k} + \pd{\mathcal{N}}{\theta_k} ) d\Omega=0.
\end{equation}

Eq. \eqref{orth2d} is satisfied when the target function is orthogonal to the image of the basis under the differential operator. The orthogonality condition we track over the course of optimization is

\begin{equation*}
    O_j(t) =  \int_0^1  \hat v(\mathbf{x}) \qty( \nabla^2 \hat h_j(\mathbf{x};\boldsymbol \theta^{\text{I}}) + \hat h_j(\mathbf{x};\boldsymbol \theta^{\text{I}})) d\Omega.
\end{equation*}

We demonstrate that exact Newton methods again find this trivial solution. The forcing function is given by $v(\mathbf{x}) = 100\sin(4\pi x_1) \sin(4 \pi x_2)$, and we use a SIREN network with two hidden layers and a width of $10$ hidden units per layer. The frequency hyperparameter is set at $\omega_0=5$. The step size and convexity parameters are $\epsilon=5 \times 10^{-2}$ and $\eta = 1 \times 10^{-1}$. The convergence threshold is set at $\mathcal{T}=1$. We also compare against ADAM optimization with a learning rate of $1 \times 10^{-2}$ with the same architecture to verify that the network is sufficiently expressive to accurately represent the true solution. We note that the exact solution to the problem can be written down by inspection as 

\begin{equation*}
    u(\mathbf{x}) = \frac{100}{32 \pi^2 } \sin(4\pi x_1) \sin(4 \pi x_2).
\end{equation*}

\begin{figure}[hbt!]
\centering
\includegraphics[width=1.0\textwidth]{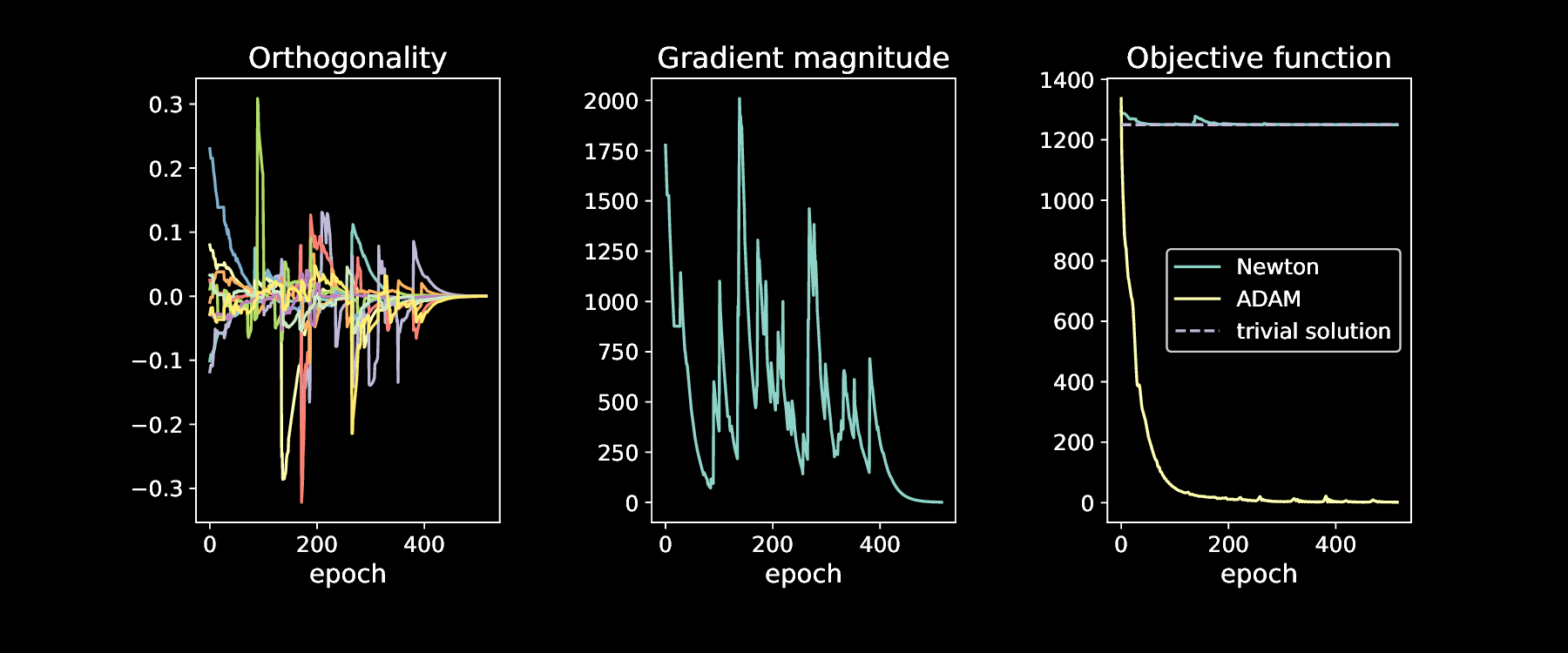}
\caption{The converged solution is such that the learned basis satisfies the differential orthogonality condition. In contrast, ADAM has no trouble finding a solution to the governing equation, indicating that the network is sufficiently expressive.}
\label{pinn_2d_conv}
\end{figure}

\begin{figure}[hbt!]
\centering
\includegraphics[width=1.0\textwidth]{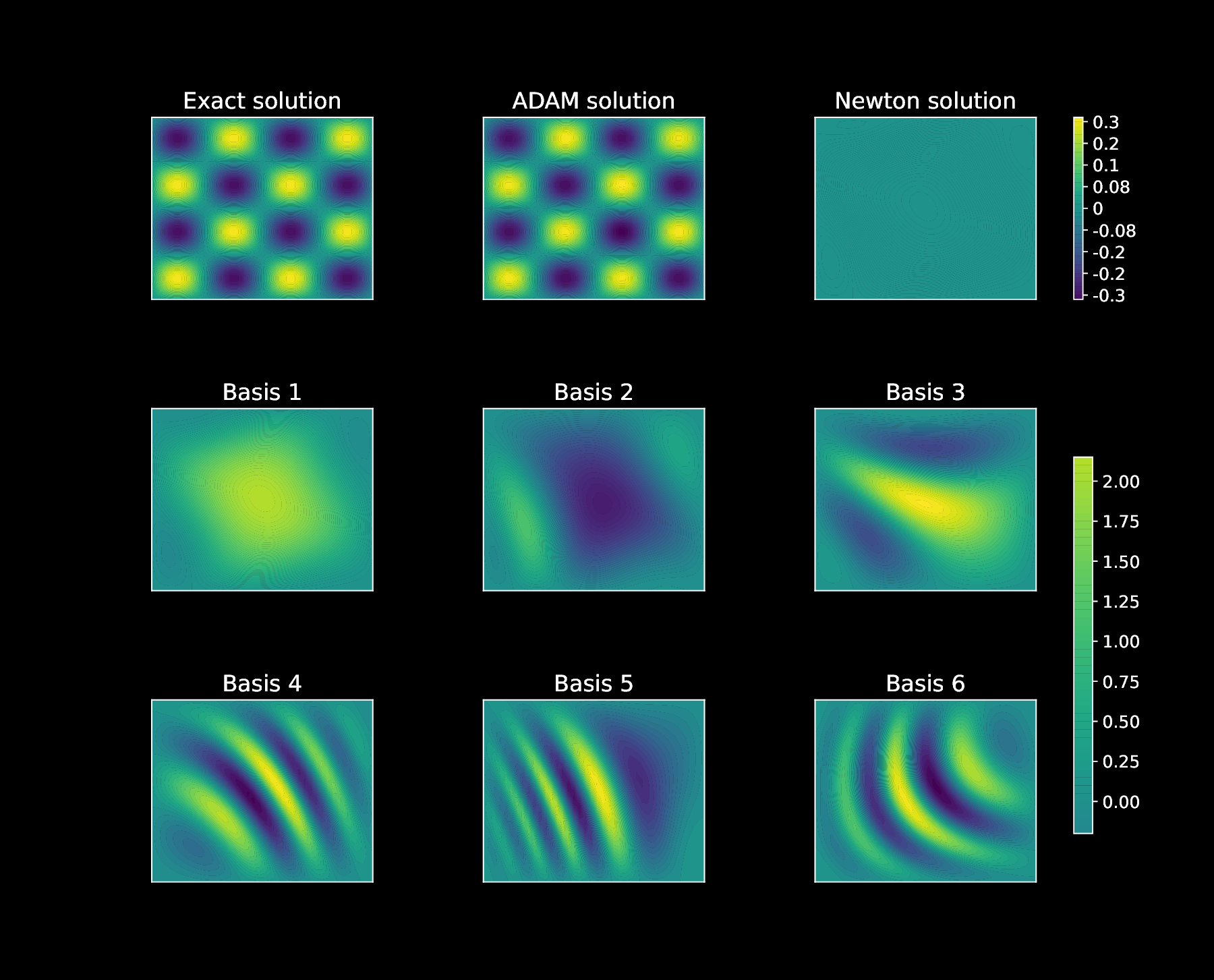}
\caption{ADAM obtains the exact solution to the PINNs problem whereas Newton finds a trivial solution. The basis functions from the SIREN network are not themselves trivial, despite the solution they ultimately construct.}
\label{pinn_basis}
\end{figure}

See Figure \ref{pinn_2d_conv} for the comparison of the Newton optimization to ADAM. As seen from the eigenvalues of the Hessian, we obtain a trivial saddle solution. This is in contrast to the ADAM optimizer, which drives the strong form loss down to near zero. Figure \ref{pinn_basis} gives a visual comparison of the ADAM and Newton solutions to the exact solution, and shows $6$ of the $10$ basis functions learned from Newton optimization. As before, it is remarkable that Newton's method finds a basis which satisfies a complex differential orthogonality condition in place of simply solving the differential equation.

\paragraph{} It is surprising that the neural network trained with Newton's method robustly finds trivial solutions, despite the complexity of the requisite orthogonality condition. We remark that we never observe the neural network finding a \textit{maximum} error solution, though this possibility cannot be ruled out. Perhaps it is the case the saddle points are more prevalent in the loss landscape than minima or maxima? While the relative frequency of the three different kinds of stationary points in high dimensions is difficult to asses theoretically, a very simple-minded argument suggests that saddle points are more common than other extrema. We know that the eigenvalues of the Hessian matrix characterize the stationary point, and we know that the Hessian matrix is symmetric. Let us assume for the sake of argument that the independent components of the Hessian are distributed as independent standard random normal variables. We assume without loss of generality that the variance of the random normal variable is unity, as a constant scale factor will not affect the sign of the eigenvalues of a matrix. The number of parameters in the one-dimensional discretizations was $|\boldsymbol \theta|=140$, so we take the Hessian matrix to be

\begin{equation}
    \mathbf{J} = \frac{1}{2}( \mathbf{M} + \mathbf{M}^T), \quad \mathbf{M} \in \mathbb{R}^{140 \times 140}, \quad M_{ij} \overset{\text{i.i.d.}}{\sim} N(0,1).
\end{equation}

We randomly generate $10^5$ such Hessian matrices and compute the eigenvalues. Out of these trials, not one random Hessian matrix has purely negative or purely positive eigenvalues. Given that the squared error regression objective is strictly positive, there is guaranteed to be a minimum somewhere, meaning there is some Hessian matrix with entirely non-negative eigenvalues. The fact of the existence of some such Hessian matrix indicates that the structure of these matrices is not purely random. Thus, the assumption of independent and random components is simply a rough model. Certainly, this model suggests that saddle points drastically outnumber minima and maxima in a high-dimensional loss landscape. This sheds some light on the reliability of the convergence to saddle solutions. The predominance of saddle points in high dimensions is discussed in \cite{dauphin_identifying_2014}, where more sophisticated arguments in favor of this conclusion are provided. That being said, we have not explained why exact Newton methods so often converge to the particular saddle point we have identified, which numerical experimentation indicates is non-unique.


\section{Conclusion: avoiding saddle points}

\begin{figure}[hbt!]
\centering
\includegraphics[width=0.75\textwidth]{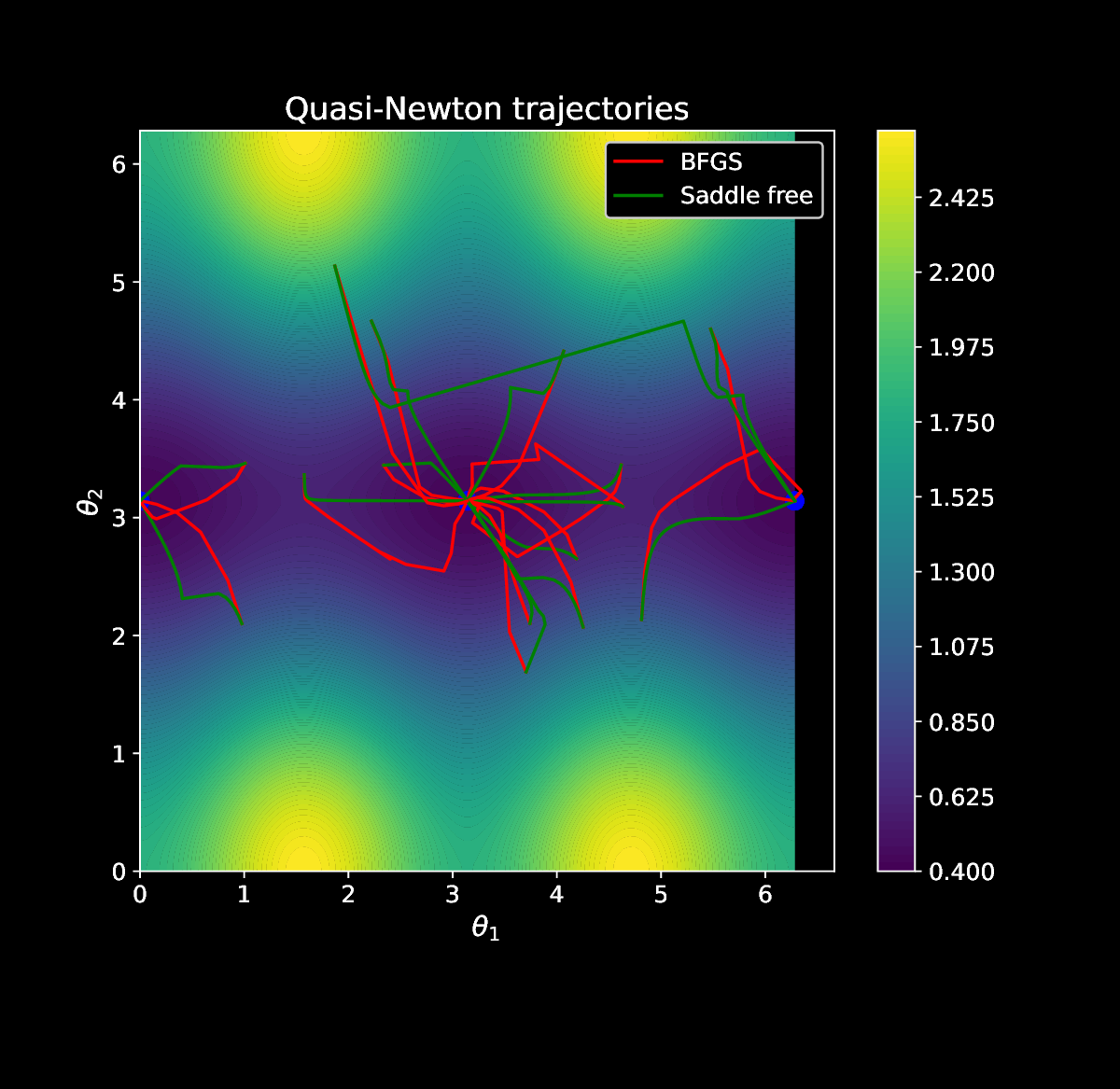}
\caption{Comparing $15$ optimization trajectories for the torus regression problem using the PyTorch BFGS quasi-Newton optimization and the saddle-free Newton method, which modifies the exact Hessian to avoid saddle points and maxima. By neglecting negative curvature, both these methods converge to one of the two minima every time.}
\label{quasi}
\end{figure}

\paragraph{} The examples given above beg the question: if Newton methods so reliably converge to trivial saddle solutions, how is it that they are used in practice to train neural networks? First, we remark that true Newton methods involving the exact Hessian are a rarity in machine learning. Though the avoidance of forming and inverting the exact Hessian is usually attributed to computational cost, our results suggest that even if the true inverse Hessian could be obtained for free, it would not be useful. Remember that Newton methods solve for a zero of the gradient of the loss, rather than explicitly minimizing the loss. Thus, given the prevalence of saddle points in loss landscapes involving neural networks, the true Hessian may point the way to nearby stationary points, rather than in descent directions. While this is well-known, we believe the following point is under-appreciated in the machine learning literature: \textit{second-order quasi-Newton optimizers succeed not in spite of their failure to approximate the true Hessian, but because of it.} These methods all have built-in safeguards against ascent in the loss landscape. BFGS and L-BFGS approximations of the Hessian enforce the so-called ``curvature condition" in order to maintain a positive definite approximation of the Hessian, even when steps in the loss landscape suggest otherwise \cite{nocedal_quasi-newton_1999}. The saddle-free Newton method modifies the true Hessian matrix in order to repel from saddle points and maxima \cite{dauphin_identifying_2014}. See \cite{simpson_regularized_2022} for a review of variants of saddle-free Newton methods. Figure \ref{quasi} depicts a return to the torus regression problem with BFGS and saddle-free Newton methods. Now, all initializations converge to one of the two minima, in spite of the prevalence of saddles and maxima in the loss surface.

\paragraph{} By explicitly minimizing the objective function, first-order optimization methods such as gradient descent and ADAM never ascend in the loss landscape to saddle solutions. They are, however, susceptible to capture by local minima and descending along ridges to saddle solutions, though such saddles are unstable fixed points of the gradient flow dynamics. We note that the standard intuition backing modifications of gradient descent such as ADAM is that the introduction of momentum allows escape from saddles and minima. If this is the case, it is not clear why one would expect a quasi-Newton optimizer---even one that avoids ascent directions---to perform better than ADAM. Under this interpretation, such a second-order optimizer---while equipped with curvature information in order to choose optimal step directions and sizes---would find the first local minimum and converge. Yet, a number of works cite quasi-Newton methods as not only being competitive with ADAM, but offering improvements in accuracy \cite{rathore_challenges_2024, kiyani_optimizing_2025}. If entrapment in local minima is a primary failure mode of an optimizer without momentum, it is not clear how to interpret the success of quasi-Newton methods. We believe there remains much intuition to acquire about the nature of loss landscapes arising from neural network discretizations and the dynamics of different optimizers.

\paragraph{} Our primary goals in this work have been to 1) to provide geometric insight into the non-convex loss landscape supplied by nonlinear discretizations, 2) characterize trivial solutions for MLP neural networks, and 3) show numerically that exact Newton methods reliably find these solutions. The robust failures of exact Newton optimization do not call into question the efficacy of second-order quasi-Newton optimization methods demonstrated in works such as \cite{urban_unveiling_2025, ahmad_preconditioned_2025, sun_physics-informed_2023}. However, our results do provide deeper insight into the nature of this efficacy. The second-order optimizer for machine learning does not improve the speed and accuracy of the solution by incorporating curvature of the loss. It does so by incorporating only that curvature information which pertains to minimizing the loss. This is the curvature which is relevant to finding descent directions and taking adaptively calibrated steps in these directions. If negative curvature information is incorporated into the Hessian, the neural network is prone to converge to a saddle solution, which our examples suggest are ubiquitous. BFGS approximations of the Hessian disregard negative curvature, whereas saddle-free Newton methods explicitly step opposite the directions of negative curvature. Though the numerical examples contained in this work are simple, we believe they are sufficient to provide worthwhile insight into the nature of loss landscapes defined by neural networks, while simultaneously surfacing fine points about the behavior of second-order optimization methods.


\end{document}